\documentclass[9pt,conference]{IEEEtran}

\usepackage{color}
\usepackage{graphicx}	
\usepackage{soul}

\usepackage{subcaption}
\usepackage{caption}
\usepackage[utf8]{inputenc}

\makeatletter
\def\hlinewd#1{%
\noalign{\ifnum0=`}\fi\hrule \@height #1 %
\futurelet\reserved@a\@xhline}
\makeatother
\usepackage{booktabs}
\usepackage{multirow}
\usepackage[binary-units=true]{siunitx}
\usepackage[algoruled,noline,longend,linesnumbered]{algorithm2e}

\usepackage{pgfplots}
\usepackage{pgfplotstable}

\usepackage{enumitem}

\usepackage{amsthm}

\begin{document}

\graphicspath{{Fig/}}
\def\figname{Figure}
\def\algname{Algorithm}

\newcommand{\papertitle}{PowerPruning: Selecting Weights and Activations for Power-Efficient Neural Network Acceleration}

\title{\papertitle}

\author{ Richard Petri$^1$, Grace Li Zhang$^2$, Yiran Chen$^3$, Ulf Schlichtmann$^1$, Bing Li$^1$\\
$^1$Technical University of Munich, $^2$Technical University of Darmstadt, $^3$Duke University\\
Email: \{richard.petri, ulf.schlichtmann, b.li\}@tum.de, grace.zhang@tu-darmstadt.de, yiran.chen@duke.edu\\
}


\IEEEoverridecommandlockouts
\IEEEpubid{\makebox[\columnwidth]{ 979-8-3503-2348-1/23/\$31.00~\copyright~2023~IEEE \hfill}
\hspace{\columnsep}\makebox[\columnwidth]{ }}




\maketitle

\begin{abstract}

Deep neural networks (DNNs) have been successfully applied in various fields.
  A major challenge of deploying DNNs, especially on edge devices, is power
  consumption, due to the large number of multiply-and-accumulate (MAC)
  operations.  To address this challenge, we propose PowerPruning, a novel
  method to reduce power consumption in digital neural network accelerators by
  selecting weights that lead to less power consumption in MAC operations.  In
  addition, the timing characteristics of the selected weights together with
 all activation transitions are evaluated. The weights and activations that lead to small delays are further selected.  Consequently, the maximum
  delay of the sensitized circuit paths in the MAC units is reduced even
  without modifying MAC units, which thus allows a flexible scaling of supply
  voltage to reduce power consumption further.  Together with retraining, the
  proposed method can reduce power consumption of DNNs on hardware by up to
  73.9\% with only a slight accuracy loss.

\end{abstract}

\section{Introduction}
\label{sec:intro} 

Deep neural networks (DNNs) have been successfully applied in various fields,
e.g., image/speech recognition.  In DNNs, a huge number of 
multiply-and-accumulate (MAC) operations with weights need to be executed, which
correspondingly causes a high power consumption in hardware. This high power
consumption poses challenges in applying DNNs on power-constrained computing
scenarios, e.g., plant disease detection in agriculture \cite{chen2020ricetalk}
and medical diagnosis devices \cite{hassantabar2022MHDeep}.  

To overcome the challenge above, various methods on software and hardware
levels have been explored.  On the software level, pruning has been proposed to
reduce the number of weights in DNNs and thus power consumption. For example,
\cite{han2016deepcompression} proposes to prune weights with small absolute
values to reduce the computation cost while maintaining inference accuracy.  In
addition, structure pruning \cite{structurepruning} 
is further developed to facilitate the mapping of DNNs onto hardware.
Besides pruning, quantization \cite{jacob2018quantization} is another major
category of methods to reduce the computation cost of DNNs. With quantization,
MAC units are implemented to process only integer instead of floating-point
arithmetic, thus leading to a significant power reduction
\cite{DBLP:journals/corr/abs-2103-13630}.

On the hardware level, various architectures have been proposed to explore how
MAC units are organized and how data flow through the accelerators to reduce
power consumption.  The systolic array from Google \cite{TPU2017,googleedgetpu}
adopts a weight-stationary data flow, where weights are stationary and
activations and partial sums are moved across the array to maximize data
reuse.  Accordingly, the amount of memory access and thus power consumption
can be reduced.  In addition, the Eyeriss structure \cite{Eyeriss2017} uses a
row-stationary data flow where the multiplication of rows of filters and
activations is computed in a MAC array to reduce data movement and thus power
consumption.

The hardware architectures above have also been extended to reduce power
consumption further.  For example, a clock-gating scheme is proposed in
\cite{effort2020} to disable the operations of unused MAC units to reduce
dynamic power consumption.  In \cite{uptpu}, power-gating unused processing
elements is proposed to reduce leakage power in idle hardware units.  In addition, an
early­stop technique in hardware has been proposed in \cite{earlystop} to skip
unnecessary MAC operations, though a complex control logic is needed to
implement this technique.  Furthermore, GreenTPU in \cite{greentpu} scales the
supply voltage of the computing logic down to near-threshold levels while
keeping a high compute performance.  But this method requires complex control
logic to detect timing errors on-the-fly and to track activation sequences that
cause timing errors.  Similarly, Minerva \cite{Minerva2016} proposes a voltage
scaling of memory units storing weights while exploiting the flexibility of
neural networks to tolerate weight errors. 

Different from the previous methods, most of which require special hardware
architecture or control logic, we propose PowerPruning, a novel method
exploiting the power and timing characteristics of weights and activations to
reduce power consumption without modifying MAC units.  PowerPruning is
the first technique to evaluate the power and timing properties of each
individual weight value and adjust neural networks accordingly.  This technique
is compatible with the previous methods for power reduction of executing neural
networks and can be integrated with them seamlessly.  The key contributions are
summarized as follows:

\begin{itemize}[topsep=3pt,itemsep=2pt,leftmargin=10pt]

 \item The power consumption of weight values is evaluated with respect to activations when the MAC
   operations are executed on hardware. Afterwards, weight values that lead to
    less power consumption in MAC operations are preferred for training
    neural networks to enhance the power efficiency. 


 \item We consider the actual delays of the MAC operations in hardware with
   respect to weight values and activations.  In training neural networks, the
    weight values and activations that sensitize paths with small delays are
    selected.  Correspondingly, the circuit can run faster without modifying MAC units. We then scale the supply voltage to reduce the power
    consumption while maintaining the original computational performance.



  \item Neural networks are retrained by restricting weights and activations to
    the selected values while maximizing the inference accuracy. With the
    selected weights and activations, power consumption of DNNs can be reduced
    by up to 73.9\% with only a slight accuracy loss.

\end{itemize}

The rest of the paper is structured as follows.
Section~\ref{sec:preliminaries} explains the motivation of this work.
Section~\ref{sec:methods} elaborates the details of the proposed technique.
Experimental results are presented in Section~\ref{sec:results} and conclusions
are drawn in Section~\ref{sec:conclusion}.

\section{Motivation}
\label{sec:preliminaries} 

In executing DNNs on hardware platforms, 
the huge number of MAC operations may consume much power.
Existing methods often introduce hardware
modifications, which may incur extra hardware cost or make the design specific
for individual neural networks. On the contrary, we address this power
consumption issue by examining the power and timing properties of the weight
values and activations. 



\begin{figure}[t]
  \begin{minipage}[c]{0.45\linewidth}
    \centering
    \includegraphics[width=0.9\linewidth]{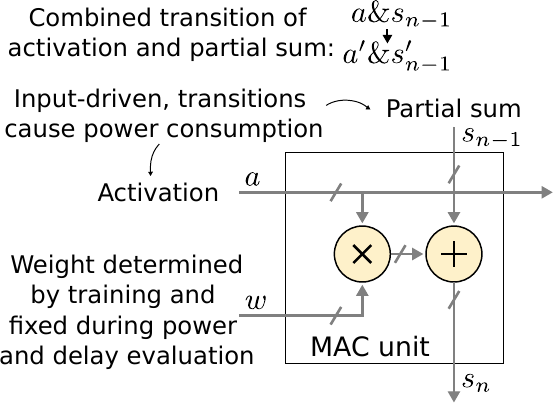}\hspace{10pt}
    \caption{Power and delay characterization of MAC unit.}
    \label{fig:mac}
  \end{minipage}\hfil
  \begin{minipage}[c]{0.45\linewidth}
    \centering
    \includegraphics[width=1\linewidth]{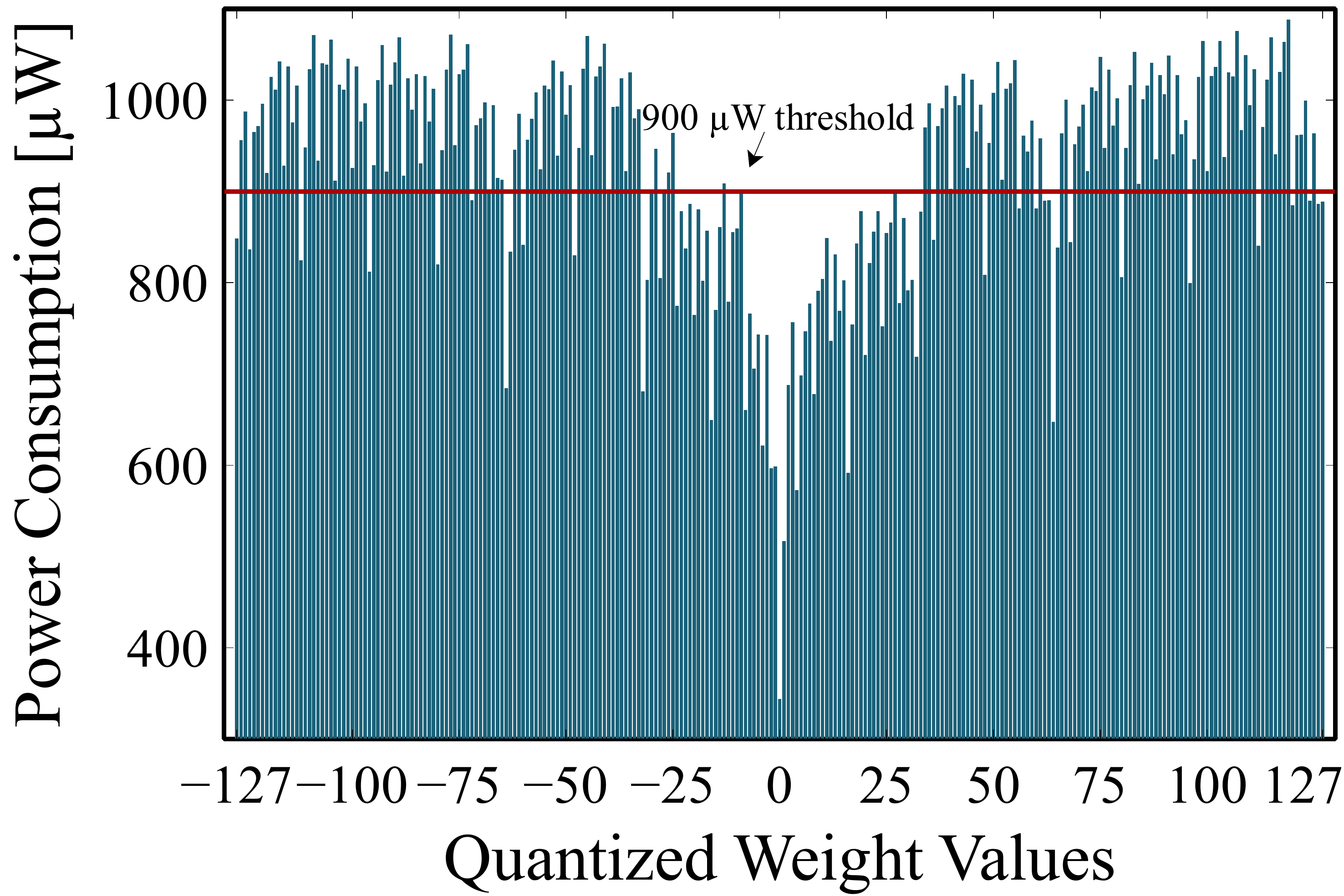}
    \caption{Average power consumption of quantized weight values.}
    \label{fig:powerResult}
  \end{minipage}
\end{figure}

A MAC unit calculates the multiplication of a weight and an activation and adds
the result to a partial sum, as illustrated in \figname~\ref{fig:mac}.  Assume
the weight of a neural network is quantized to $n$ bits.  Correspondingly,
there are $2^n$ possible weight values.  These weight values are one of the inputs to the
digital logic implementing the MAC operations.  Since different weight values
cause different signal switching activities inside the MAC units, they also exhibit
different average power consumption with respect to the activation transitions and partial sum transitions. For
example, the weight values $2^n, n=0,1,\dots,n-2$, lead to less power
consumption, because the multiplication with these weight values are actually
shift operations and can thus activate fewer signal propagations in the
circuit.

To demonstrate the different power consumption of weight values, we evaluated
the average power consumption of different weight values in a MAC unit of a
$64\times64$ systolic array. We simulated the execution of LeNet-5 processing
100 pictures randomly selected from the CIFAR-10 dataset.
During simulation, we collected statistics of the switching activities of various signals inside the systolic array.
Based on this data we estimated the average power consumption of each weight value using Power Compiler from Synopsys.



Figure~\ref{fig:powerResult} illustrates the average power consumption of the
weight values obtained by the simulation described above.  According to this
figure, different weight values can lead to substantially different
average power consumption.  For example, the quantized weight value -105 has a large
average power consumption 1,066\,$\mu$W, while the quantized weight value -2 has only
596\,$\mu$W.  \textit{According to this observation, by restricting neural
networks to prefer the weight values with small average power consumption, the
overall power consumption of executing neural networks can be lowered.} 


Besides different power characteristics, different weights also exhibit
different timing profiles in a MAC unit.  Inside a MAC unit shown in
\figname~\ref{fig:mac}, there are many combinational paths, which have different
delays and are triggered by specific input data, i.e., weight, activation, and
partial sum.
If the weight is fixed to a given value, some combinational paths in the MAC
unit cannot be sensitized.  Accordingly, the delay of the MAC unit may differ
with respect to different weight values.  To demonstrate this difference, we
conducted timing analysis of the MAC unit with fixed weight values and all
activation transitions using Modelsim.

\begin{figure}[t]
    \centering
    \includegraphics[width=0.95\linewidth]{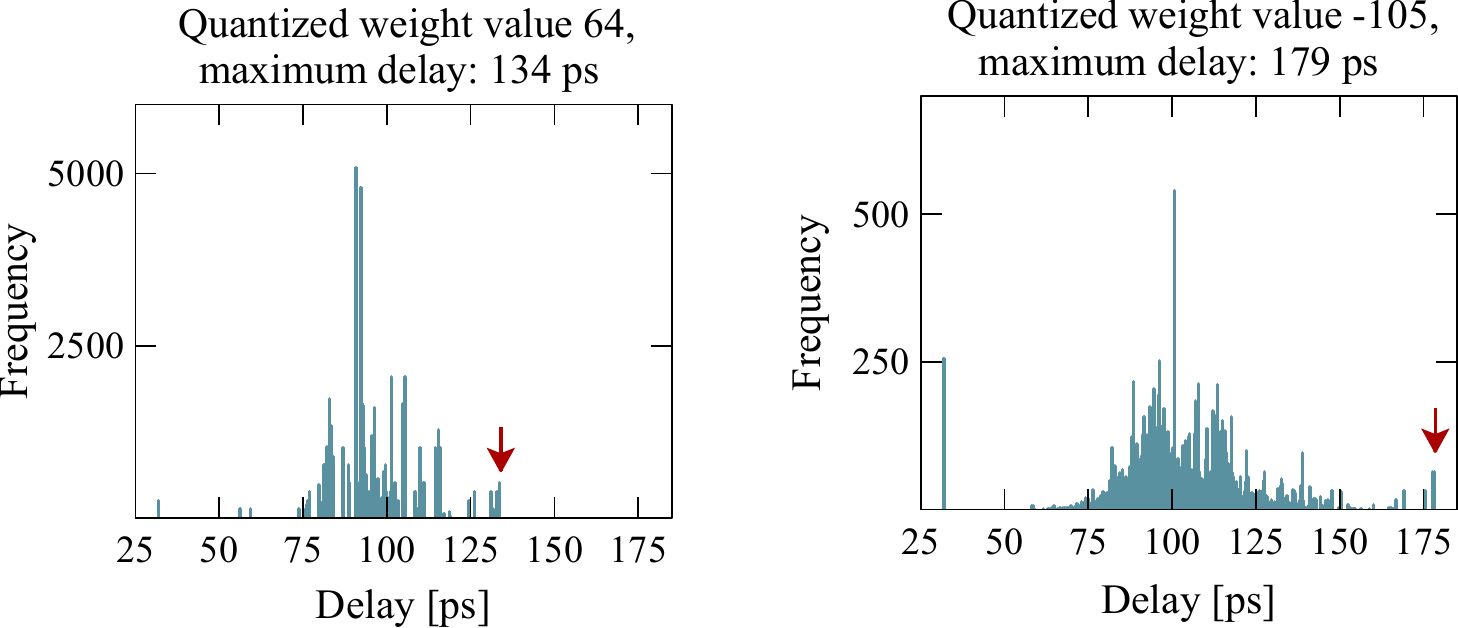} 
    \caption{Delay profiles of a MAC unit for two quantized weight values.
    The arrows point to the maximum delay of a given weight value with respect
    to all the activation transitions.}
    \label{fig:delayResult}
\end{figure}

Figure~\ref{fig:delayResult} illustrates the delay profiles of two quantized weight
values -105 and 64, where the x-axis shows the delay and the y-axis shows the
frequency of this delay appearing with respect to all possible activation transitions.
Figure~\ref{fig:delayResult} confirms that different weight
values lead to different delays. In addition, it shows that the
delays can be reduced further if some activations can be
pruned from the neural network, e.g., the activation transitions triggering
delays on the far right end of the x-axis.  \textit{Since the clock period of
a circuit is determined by the maximum delay of all the combinational paths,
the clock frequency of the MAC unit and thus the computational performance can
be increased by
pruning weights and activations according to their timing profiles.  Alternatively, the supply voltage can be lowered
to reduce power consumption further, while maintaining the original clock
frequency.}

\section{Weight and Activation Selection for Power-Efficient Neural Network Acceleration}
\label{sec:methods}

In this section, we introduce the proposed PowerPruning method to reduce power
consumption in digital neural network accelerators.  The weight selection
according to the average power consumption is first explained in
Section~\ref{sec:power}.  Afterwards, the selection of weights and activations
with respect to their timing characteristics is explained in Section~\ref{sec:timing}.  The retraining of
neural networks by restricting weights and activations to the selected values to reduce power consumption  
is described in Section~\ref{sec:training}. 

\subsection{Weight selection according to power consumption}\label{sec:power}

As shown in Figure~\ref{fig:powerResult}, different weights in a MAC unit lead
to different average power consumption.  To take advantage of this
characteristic to reduce power consumption of DNN accelerators, the average
power consumption of all the 8-bit integer weight values in a MAC unit should
be evaluated. To do this, the input of the MAC unit corresponding to the weight
is fixed to a given value, as shown in \figname~\ref{fig:mac}. The various
combinations of activation transitions and partial sum transitions are fed into
the other inputs of the MAC unit to obtain the switching activities of the MAC
unit.  Based on these switching activities the power consumption for the fixed
weight value can be evaluated using  Power Compiler from Synopsys. 

Two challenges in evaluating the average power consumption of a weight should
be addressed. First, the number of combined transitions of activations and
partial sums is huge, e.g., $2^{(8+22)\times 2}=2^{60}\approx10^{18}$, when the
activations and the partial sums are quantized to 8 and 22
bits, respectively, for a $64 \times 64$ systolic array.  $\times 2$ is due to
the fact that the power consumption is caused by the transitions from a
combination of activation and partial sum to another combination, instead of
the static values of the activation and partial sum.  Accordingly, simulating
all these transitions to identify the power consumption is very time-consuming.
Second, just sampling all possible combined transitions of activations and
partial sums does not reflect the probabilities of such transitions when
executing neural networks in a systolic array.  For example, a combined
transition may appear more frequently than other transitions, so that it should
contribute more to the result of power evaluation than others.

To deal with these challenges, 
we first identify the transition distributions for activations 
and partial sums with real data executing on the systolic array, described as
follows. In addition, we partition the value range of the partial sum into a small number of bins to reduce the partial sum transition space and then evaluate the transition probability from one bin to another bin.


\subsubsection{Evaluation of activation transition distribution}
\label{sec:evac}

For the 8-bit activation as an input to a MAC unit, the total number of possible
transitions is $2^{8\times 2}=2^{16}$.  To obtain the activation transition
distribution, we simulate the activities of a systolic array 
and count the frequency of each individual transition.  For example, for
LeNet-5 on CIFAR10, we randomly select 100 pictures and execute the neural
network on the systolic array. In total we  counted approximately $10^{17}$
activation transitions.  Since this number is larger than the  number of
possible transitions $2^{16}$, the result will well exhibit the distribution of
the activation transitions. 

\begin{figure}
     \centering
     \hfil
     \begin{subfigure}[b]{0.45\linewidth}
         \centering
         \includegraphics[width=0.72\linewidth]{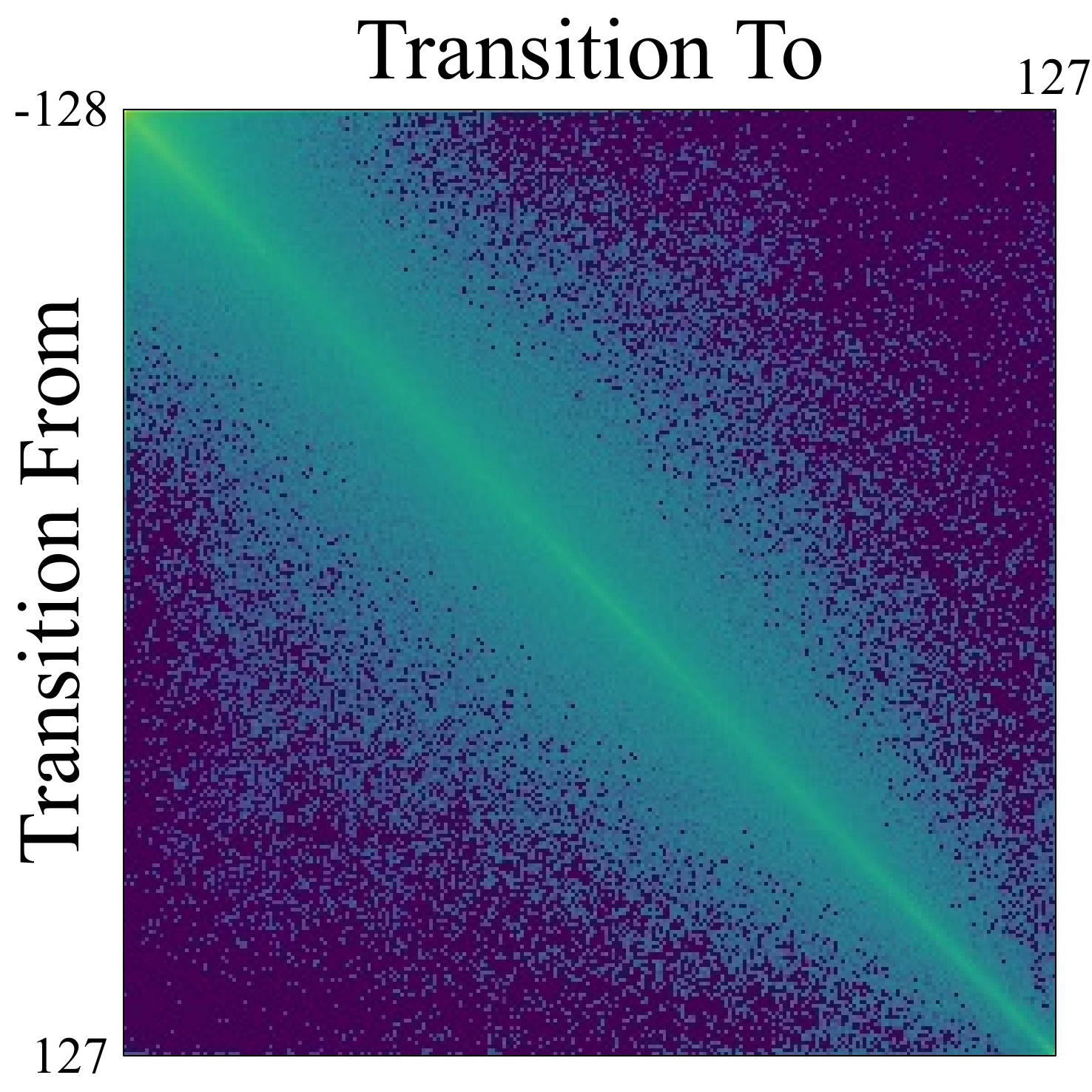}
         \caption{}
         \label{fig:actDis}
     \end{subfigure}
     \begin{subfigure}[b]{0.45\linewidth}
         \centering
         \includegraphics[width=0.72\linewidth]{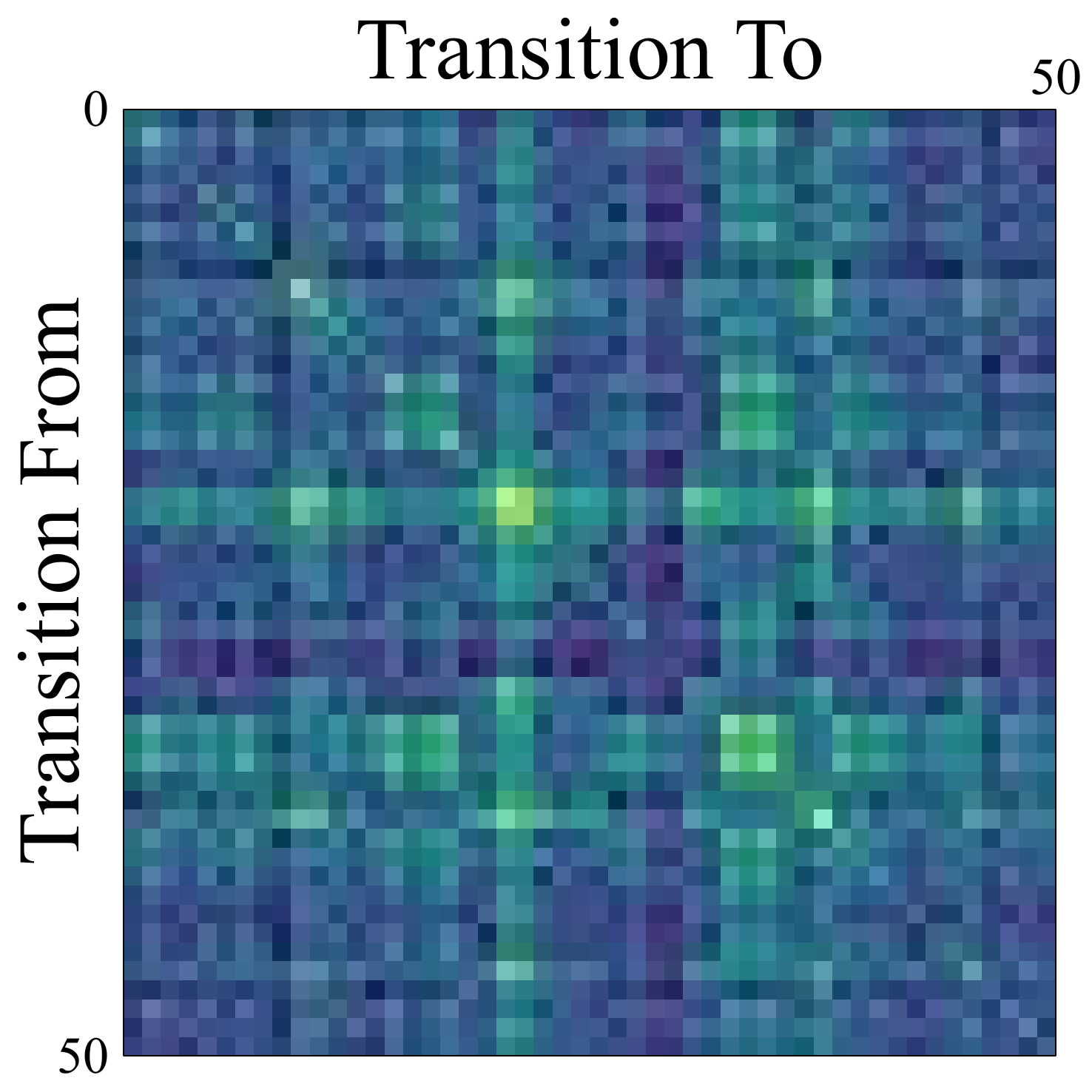}
         \caption{}
         \label{fig:PartSumDis}
     \end{subfigure}
     \hfil
        \caption{Transition distributions of activations and partial sums of a MAC unit.
  (a) Activation transition distribution. (b) Partial sum transition distribution.}
        \label{fig:trans}
\end{figure}

Figure~\ref{fig:trans}(a) shows the resulting activation transition distribution, where 
darker colors represent a lower probability and brighter colors a higher probability. 
In this figure, the bright diagonal line clearly indicates that most
transitions appear between activations with similar values, while activation
transitions from very high to very low values and vice versa are very unlikely
to happen.

\subsubsection{Evaluation of partial sum transition distribution and transition
space reduction}
 A partial sum has 22 bits in a systolic array with the size of $64 \times 64$,
which results in $2^{22\times 2}=2^{44}\approx1.8\times 10^{13}$ possible
transitions. If we would simulate 100 pictures on the systolic array,  we can
obtain approximately $2.2\times 10^{8}$ partial sum transitions, which is much
smaller than the number of possible transitions and cannot produce a
trustworthy distribution.  Increasing the number of pictures in simulation is
not a viable solution due to runtime.  To solve this problem, we partition the
value range of the partial sum into a small number of bins.  Accordingly,
instead of evaluating the transition probability of individual partial sum
values, we evaluate the transition probability from one bin to another bin.

To partition all partial sums into a small number of bins, 
the switching activities of partial sums from one bin to another bin should be maintained as similar as possible. 
To achieve this goal,
we group partial sums 
according to the similarity in their bit numbers.
Specifically, a predetermined number of bins is first specified. Afterwards, partial sums are randomly selected and assigned into each bin. 
The remaining partial sums are iteratively assigned into the most matching bins by
measuring the similarity between a specific partial sum and a bin. This similarity is evaluated 
by counting the number of different bits between this partial sum and those in the bins. 
The partial sum is assigned to
the bin where the partial sum has the least number of different bits in average. 
In the experiments, 50 bins were used.



After the partition of partial sums into bins, we simulated 100 pictures and
assigned the real transitions into these bins. Afterwards, the probabilities of
the transitions between bins can be identified, similar to the evaluation of the
activation transition distribution in Section~\ref{sec:evac}.
Figure~\ref{fig:trans}(b) shows the partial sum transition distribution of
the bins. It can be observed that 
the partial sum transitions are not evenly distributed.  
For example, the bright diagonal line from the upper left corner to the lower right corner indicates that there are many transitions between partial sums with similar values. In addition, the bright vertical and horizontal lines also demonstrate intensive partial sum transitions. 


\subsubsection{Weight selection}

With the distributions identified above, we sample 10,000 transitions of both
activations and partial sums according to their probabilities. The combined
transitions are used to simulate the activities of the MAC unit with the weight
input fixed to specific values.  The resulting switching activities are then
used to calculate the average power consumption of the MAC unit for this
weight. This simulation is repeated for each individual weight value and the
result is shown in \figname~\ref{fig:powerResult}, where the power consumption
of each weight varies greatly.  In this result, there is also a trend that
weights close to zero have especially low power consumption, with weight zero
having by far the lowest.

Based on the result of power analysis we first conduct conventional pruning to
maximize the number of weights with zero value to reduce power consumption.  Afterwards, we
select weight values that lead to small power consumption by setting a power
threshold, e.g., 900\,$\mu$W in \figname~\ref{fig:powerResult}.  By setting the
threshold lower, we can achieve potentially more power savings by excluding
more high-power weight values.  However, the accuracy of the DNN may degrade.
Therefore, a tradeoff between power saving and inference accuracy should be
made.

\subsection{Weight and activation selection according to timing profiles}\label{sec:timing}

According to Figure~\ref{fig:delayResult}, weight values exhibit different
timing characteristics.  Even for the same weight, different activation
transitions lead to different delays.  To identify weight values and
activations with small delays,
the timing of each weight value with respect to activation transitions and
partial sum transitions in the MAC unit should be analyzed.  Two types of
timing analysis methods, dynamic timing analysis and static timing analysis,
are available for this task.  The former is conducted by applying input
transitions into a circuit and evaluating the delays of correspondingly
triggered paths, while the latter evaluates the delay statically without
considering the corresponding triggered paths.  The latter is conservative
since the delays of some paths that are not activated are also included and the
clock frequency of the circuit may be unnecessarily lowered.

To evaluate the timing profile of a weight value, an intuitive idea is to fix
the weight input into the MAC unit and then apply dynamic timing analysis with
all the transition patterns of activations and partial sums to simulate the
unit.  The challenge of this method is that the number of combined transitions
of activations and partial sums is huge, as described in
Section~\ref{sec:power}.  Simulating the delay of the MAC unit with respect to
all these combinations is thus time-consuming. 

To reduce the runtime of timing analysis, we separate the timing analysis of the
multiplier and adder in the MAC unit.  Specifically,  we apply static timing
analysis on the adder to avoid the consideration of input
transitions, because the number of inputs to the adder is very large.  On the
other hand, the multiplier is evaluated using accurate dynamic timing analysis,
since the delay of the multiplier usually dominates the delay of the MAC unit
and this delay can be lowered by filtering out some weight values and
activation values.

To conduct dynamic timing analysis of the multiplier for a weight value, we
simulate the multiplier by fixing the weight input and enumerating the
$2^{8\times 2}$
possible transitions of the activations. Static timing analysis of the adder is
conducted by the built-in timing analyzer in Design Compiler from Synopsys.
To incorporate the relation between the timing paths in the multiplier and the
adder, 
we evaluate the largest delay starting from each individual bit of the product
to the output of the adder.  Afterwards, the largest delay of the MAC unit with
respect to the given weight value is calculated by adding the delays from the input activation to the
output bits of the multiplier and the  delays from the corresonding
product bits to the output of the adder.

\begin{figure}[t]
    \centering
    \includegraphics[width=0.7\linewidth]{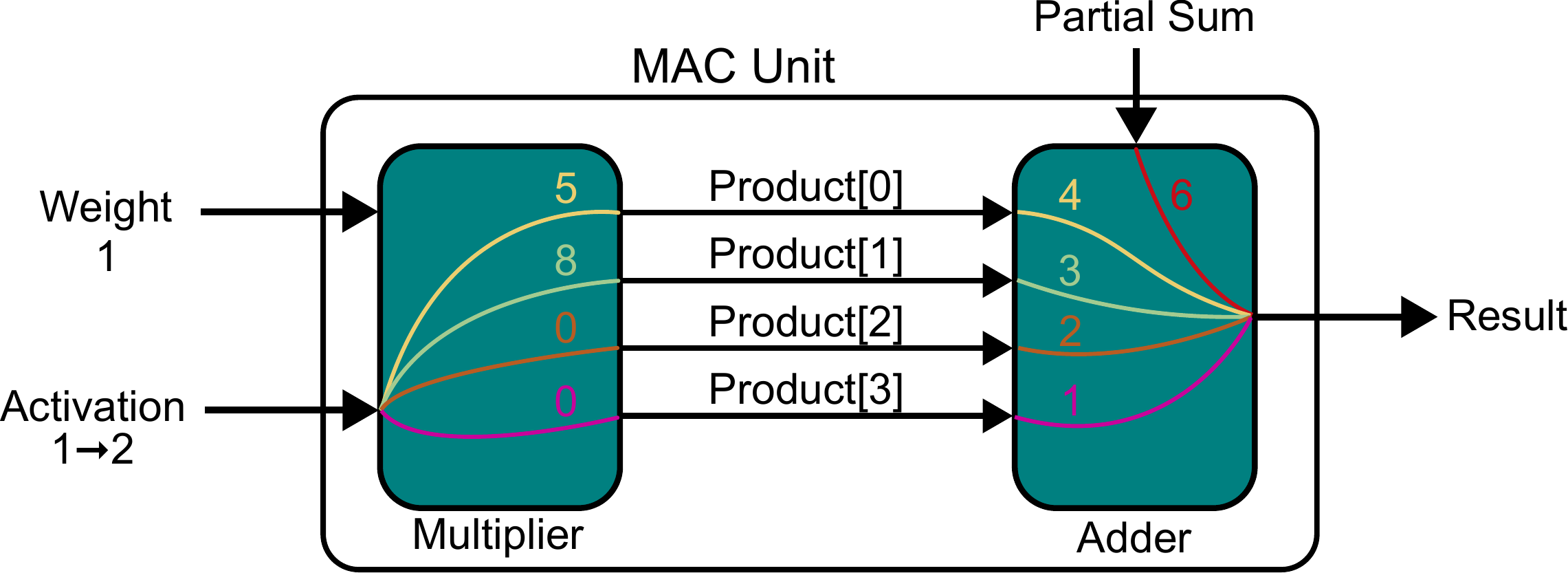}
    \caption{Concept of timing analysis of the MAC unit.} 
    \label{fig:timingMac}
\end{figure}

\begin{figure}[t]
    \centering
    \includegraphics[width=0.7\linewidth]{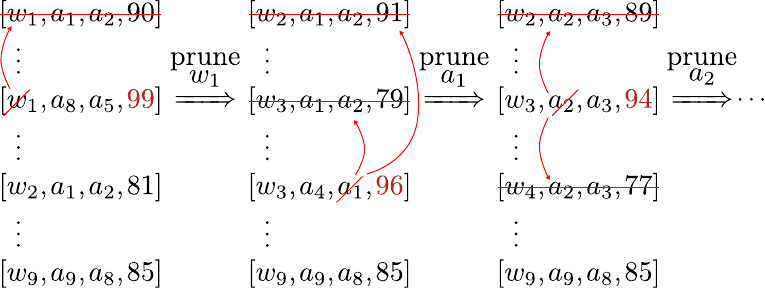}
    \caption{Concept of weight and activation selection for delay reduction.}
    \label{fig:delayOptimization}
\end{figure}
Figure~\ref{fig:timingMac} illustrates the concept of timing analysis of the
MAC unit, where the quantized weight 1, the activation transition from quantized 1 to quantized 2,
and four product bits at the output of the multiplier are used as example.
With the dynamic timing analysis applied on the multiplier, the delays from the
input activation to Product[0] and Product[1] are 5 and 8, respectively.  The
delays to the other two product bits can be 0 if the combinational paths to
them are not activated by the activation transition.  With static timing analysis applied
on the adder, the delays from the output bits of the multiplier to the output
of the adder are 4, 3, 2, 1, respectively.  Assume the delay from the partial sum
to the output of the adder is 6 returned by static timing analysis.  The
largest delay of the MAC unit is thus $\max\{5+4, 8+3, 6\}=11$. 


The timing analysis method described above is applied for each weight
individually.  After that, all the delays of weights with respect to activation
transitions can be obtained.  To select weights and activations with small
delays, we first set a delay threshold and iteratively remove weights or
activations that lead to delays larger than the given threshold.  The
iterations end until all the delays of the remaining weights and activations
are smaller than the given delay threshold.  Figure~\ref{fig:delayOptimization}
illustrates an example with the delay threshold set to 90.  In the first step,
we find the largest delay 99.  Since 99 is larger than the specified threshold,
we have to remove either $w_{1}$, $a_{5}$ or $a_{8}$ to exclude the
correponding combination. Since the removal of either $w_{1}$, $a_{5}$ or
$a_{8}$ also affects other combinations in
\figname~\ref{fig:delayOptimization}, it is difficult to find the optimal
sequence to remove the weigths and activaitons. Accordingly, we randomly remove
any of them and then remove the other combinations containing the removed
weight or activation in \figname~\ref{fig:delayOptimization}.  For example,
removing $w_{1}$ also leads to the removal of the first combination in
\figname~\ref{fig:delayOptimization}. To avoid local optimum, we execute this
process several times and choose which weight or
activation to remove in each step randomly.  The removal process ends when the maximum
delay of all the combinations is lower than the given threshold 90.  The result
is a set of weights and activations that satisfy the delay requirements. While
pruned weight values can be avoided during training of neural networks, the
filtering of activations needs to be integrated into the activation function
after each layer.




\subsection{Neural network training for power reduction}\label{sec:training}



\setlength{\tabcolsep}{1.5pt}
\renewcommand{\arraystretch}{1.2}
\begin{table*}[t]
\centering
\caption{Experimental results of proposed method.}
\begin{tabular}{lcccccccccccccccclccccc}
\hline
                              &  &               &              &  & \multicolumn{7}{c}{Total   Power Consumption {[}mW{]}}                &  & \multicolumn{2}{c}{}          &  &                                                                           &  & \multicolumn{1}{l}{}                                                            \\ \cline{6-12}
                              &  & \multicolumn{2}{c}{Accuracy} &  &
			      \multicolumn{3}{c}{Standard HW} &  &
			      \multicolumn{3}{c}{Optimized HW} &  &
			      \multicolumn{2}{c}{\#Selected} &  &
			      \multirow{2}{*}{\begin{tabular}[c]{@{}c@{}}Max
				Delay\\ Red.\end{tabular}} &  &
				\multirow{2}{*}{\begin{tabular}[c]{@{}c@{}}Voltage
				  Scaling\\ Factor\end{tabular}} &
				  \multirow{2}{*}{\begin{tabular}[c]{@{}c@{}}
				    \\ V\_SHW\end{tabular}} &
				    \multirow{2}{*}{\begin{tabular}[c]{@{}c@{}}\\
				    V\_OHW\end{tabular}}\\ \cline{3-4} \cline{6-8} \cline{10-12} \cline{14-15}
Network-Dataset               &  & Orig.         & Prop.        &  & Orig.   & Prop.     & Red.      &  & Orig.   & Prop.     & Red.      &  & Wei.            & Act.          &  &                                                                           &  &                                                                                 \\ \hline
LeNet-5-CIFAR-10              &  & 80.7\%        & 78.4\%       &  & 281.6   & 152.1   & 46.0\%  &  & 280.4    & 73.1    & 73.9\%  &  & 32            & 176           &  & 40 ps                                                                    &  & 0.71/0.8  & 13.7\%  & 6.4\%                                                                       \\
ResNet-20-CIFAR-10            &  & 91.9\%        & 88.9\%       &  & 469.9   & 230.6   & 50.9\%  &  & 427.7    & 173.4   & 59.4\%  &  & 32            & 176           &  & 40 ps                                                                    &  & 0.71/0.8    & 12.7\%  & 10.6\%                                                                     \\
ResNet-50-CIFAR-100           &  & 79.9\%        & 78.4\%       &  & 509.1   & 278.7   & 45.3\%  &  & 510.8    & 140.8   & 72.4\%  &  & 40            & 220           &  & 30 ps                                                                    &  & 0.73/0.8  & 10.6\%  & 5.2\%                                                                      \\
EfficientNet-B0-Lite-ImageNet &  & 74.4\%        & 72.9\%       &  & 152.0    & 106.7    & 29.8\%   &  & 134.2      & 78.5     & 41.5\%  &  & 76            & 244           &  & 20 ps                                                                    &  & 0.75/0.8  & 8.8\%  & 8.0\%                                                                      \\ \hline
\end{tabular}
\label{tab:results}
\end{table*}

To reduce power consumption of DNNs on hardware, we first apply conventional
pruning to remove weights whose absolute values are close to zero.
Afterwards, we select weights that lead to small power
consumption by setting a power threshold. The initial power threshold is
900\,$\mu$W and it is iteratively reduced to select weights.  In each
iteration, the neural networks are retrained with the selected weights to
verify the inference accuracy.  During retraining, we force the
weights to take the restricted values in the forward propagation. In the backward
propagation, the straight through estimator \cite{Bengio2013EstimatingOP} is
adopted to skip the restriction operation. The iterations end when the
inference accuracy starts to drop noticeably.

After the power threshold is determined, we then select weight values and
activations that lead to small delays by setting a delay threshold.  The
initial delay threshold is 170\,$ps$ and the delay
threshold is iteratively reduced by 10\,$ps$ to select weight values and
activations. In each iteration, the neural networks are retrained and verified.
When the inference accuracy drops by around 5\% of the original inference
accuracy of the neural networks, the best training result is returned.

When executing the neural networks, if the original clock frequency should be
maintained, we can lower the supply voltage to reduce power reduction. We  
use the results in \cite{power2014} to determine the relation between
supply voltage and the delay of the circuit. The scaling of dynamic power
consumption and leakage is conducted according to \cite{7747444}.

\section{Experimental Results}
\label{sec:results}

To verify the proposed method, we tested four different neural network and
dataset combinations, as shown in the first column of Table~\ref{tab:results}.
The weights and activations were quantized to eight bits.  The neural networks
were trained using Tensorflow while considering quantization
\cite{jacob2018quantization}. In Tensorflow, the number of 8-bit weights is 255 instead of 256 to maintain the weight distribution symmetrical while the number of 8-bit activations is 256. After training, small weights were pruned to compress the
neural network. We then applied the proposed method to reduce the
power consumption.  For LeNet-5, ResNet-20 and ResNet-50, Nvidia Quadro
RTX 6000 GPU 24 GB was used for training, and for EfficientNet-B0-Lite Nvidia
A100 80 GB GPU was used. The number of times to execute the selection of weight and activation with small delays in Section~\ref{sec:timing} is set to 20 in the experiments. 





To demonstrate the effectiveness of the proposed method on different types of
accelerators, two different hardware implementations of systolic array were
evaluated.  In the optimized hardware architecture (Optimized HW), clock gating
of a MAC unit in case of a zero weight to reduce dynamic power
consumption and power gating of whole unutilized columns in the systolic array
to reduce both the dynamic and static power consumption are
applied.  In the standard architecture (Standard HW), none of these
power-saving features were applied.

The power consumption during inference was estimated with Power Compiler by simulating the systolic
array executing the neural networks using Modelsim. 
The simulation was conducted using a netlist description of the systolic array synthesized with
the NanGate 15\,nm cell libraries \cite{Nangate} and the clock frequency around
5\,GHz.  Since cycle-accurate simulations are extremely time-consuming, for
ResNet-20, ResNet-50 and EfficientNet-B0-Lite only the convolutional layers
with the largest number of MAC operations were simulated and compared.

\begin{figure}
    \centering
         \includegraphics[width=0.95\linewidth]{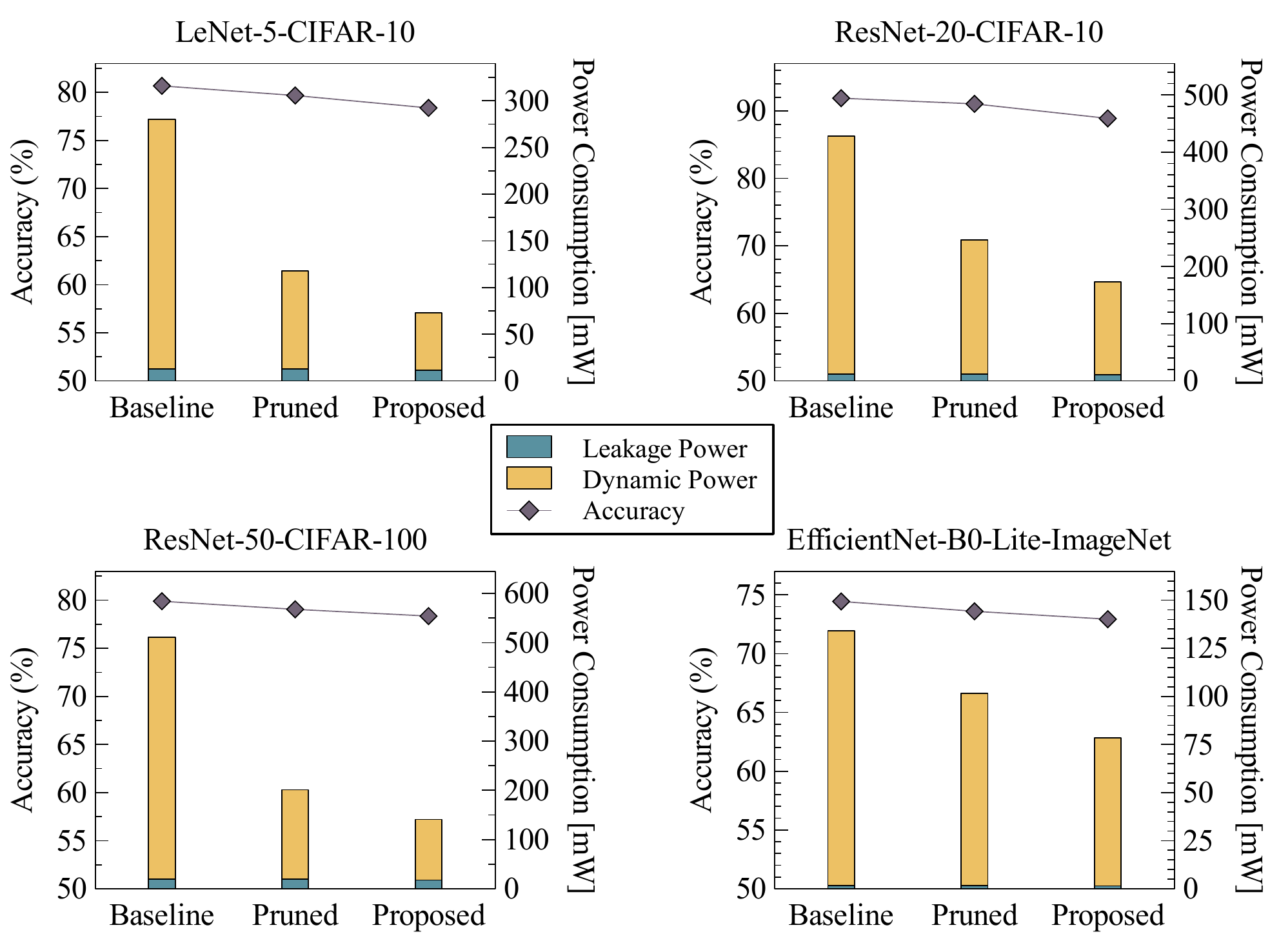}
         \caption{Comparison with conventional pruning, evaluated on Optimized HW.}
         \label{fig:overallTradeoff}
\end{figure}

Table~\ref{tab:results} summarizes the experimental results of the proposed
method. The original accuracy and the accuracy with our method are shown in
the second and third columns.  According to these two columns, the accuracy
degradation is relatively small,
With a slight accuracy loss, a significant reduction in power consumption
up to 73.9\% can be achieved, as shown in the sixth and ninth columns,
demonstrating the effectiveness of the proposed method in enhancing power
efficiency of digital accelerators for neural networks. This is especially
useful for edge devices where power consumption is a major issue. 

When executing the neural networks on Standard HW, the total power including
dynamic and leakage power was reduced by up to 50.9\% (sixth column).  On
Optimized HW, the power saving was even greater with a power reduction up to
73.9\% (ninth column). The relatively smaller power reduction on
Standard HW was caused by the leakage power consumption from the MAC units that
were not gated even when they are not used.





To reduce the maximum delay of a MAC unit, which was 180\,ps after synthesis,
we only selected a subset of weight values and activations that lead to small
delays of the MAC operations.  The number of selected weight values and
selected activation values are shown in the tenth column (Wei.) and the
eleventh column (Act.).  According to the tenth column, the number of selected
weight values is reduced significantly, e.g., from 255 to 32 in LeNet5 and
ResNet-20. On the contrary, most activation values still remain to maintain a
good inference accuracy.  The delay reduction due to the weight and activation
selection is shown in the twelfth column (Max Delay Red.). In identifying delay
reduction, our search granularity was 10\,ps. This can be lowered if necessary,
but at the expense of more runtime.


To reduce power consumption further, the supply voltage is lowered by the ratio
shown in thirteenth column (Voltage Scaling Factor). For example, for
LeNet-5-CIFAR-10, the supply voltage was reduced from 0.8\,V to 0.71\,V while
still maintaining the original clock frequency.  The relation between supply
voltage scaling and circuit delay was evaluated according to the simulation
results in \cite{power2014}. The last two columns show the percentage of power
reduction contributed by voltage scaling. For Standard HW (column V\_SHW) and
Optimized HW (column V\_OHW), voltage scaling can reduce power consumption by
up to 13.7\% and 10.6\%, respectively. 

To demonstrate the advantage of the proposed method over conventional pruning,
we show the comparison of the power consumption and the inference accuracy of
conventional pruning and the proposed method in
Figure~\ref{fig:overallTradeoff}.  According to this comparison, the proposed
method can significantly reduce the power consumption of a pruned neural
network further with only a slight accuracy loss.



\begin{figure}
         \centering
         \includegraphics[width=0.95\linewidth]{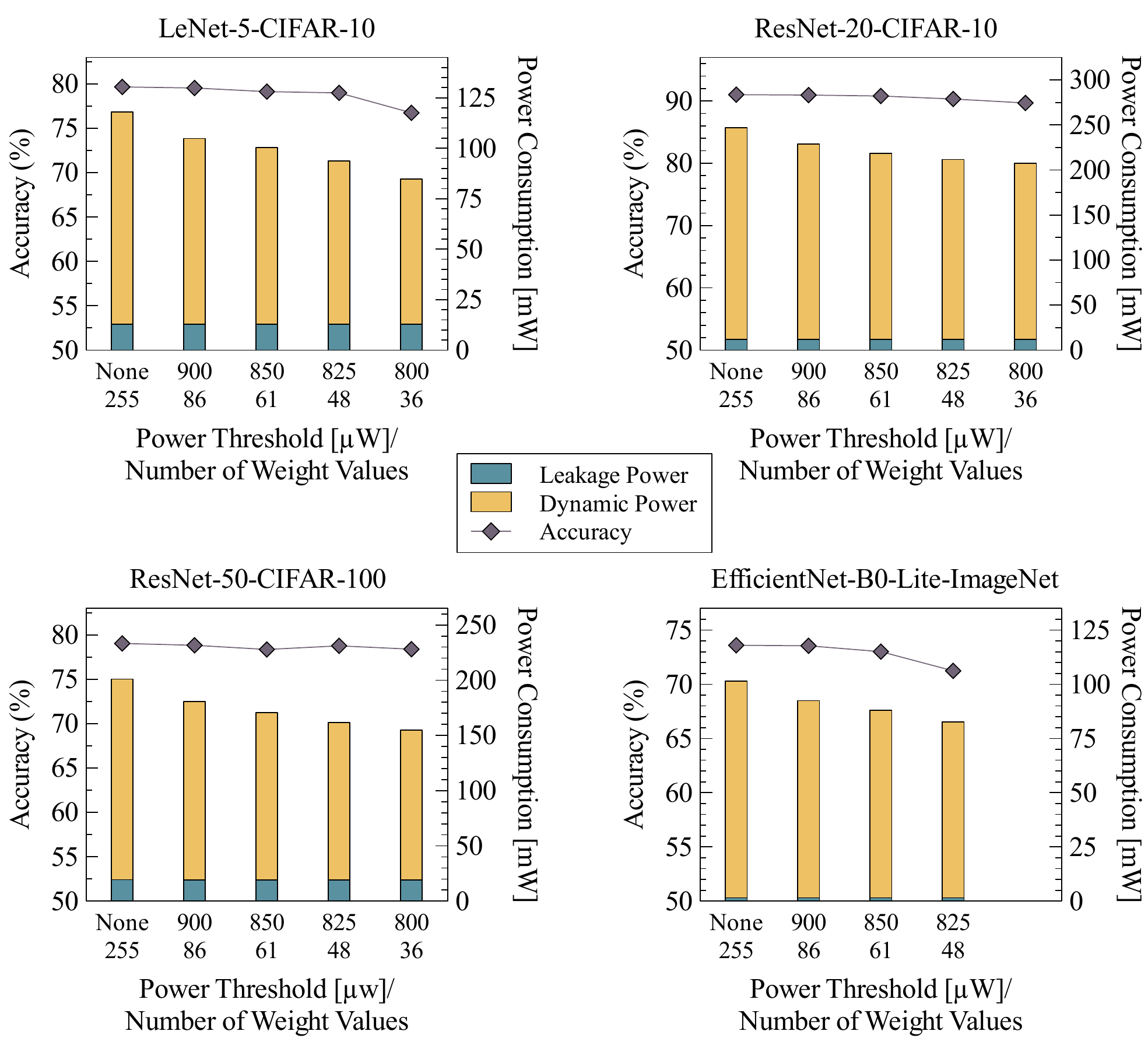}
         \caption{Tradeoff between accuracy and the number of selected weight
	 values, evaluated on Optimized HW.}
         \label{fig:spaceRestriction}
\end{figure}

\begin{figure}
         \centering
         \includegraphics[width=0.95\linewidth]{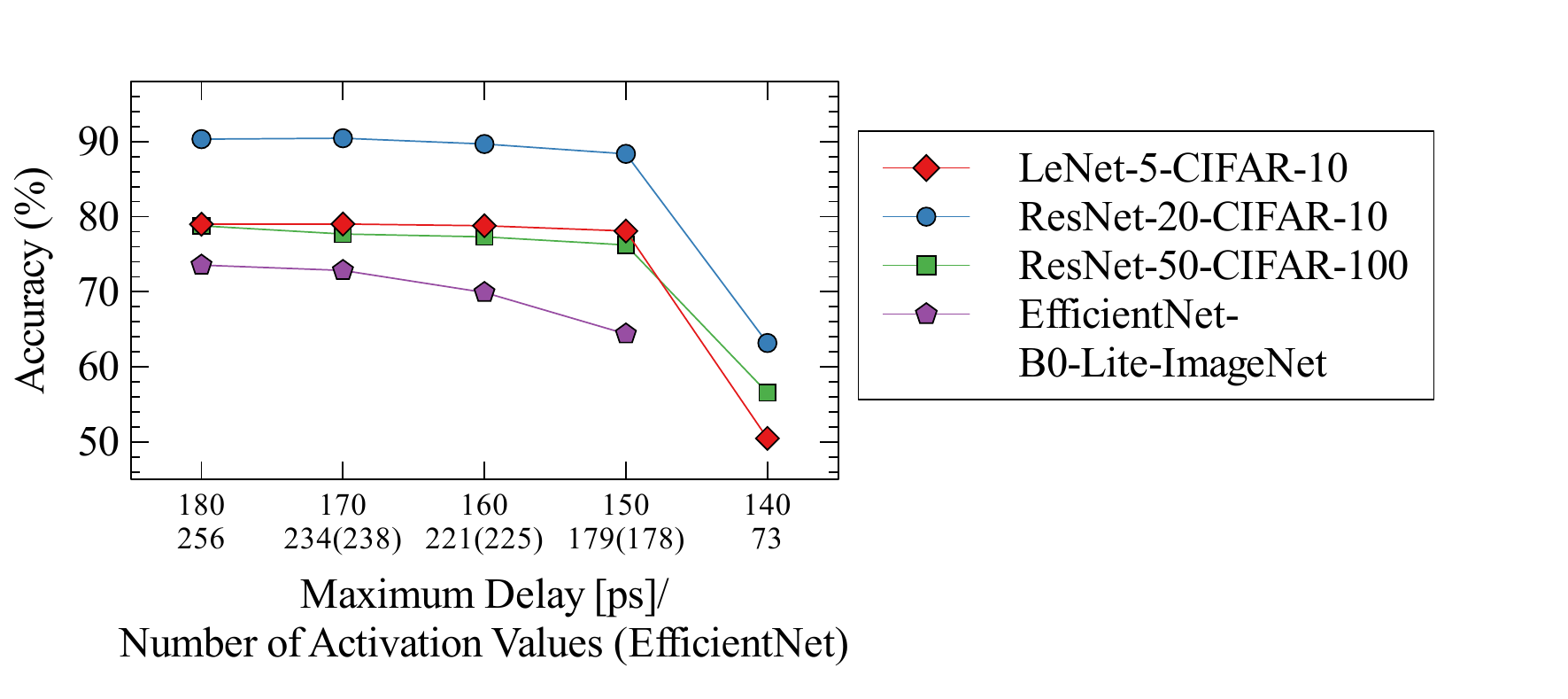}
         \caption{Tradeoff between accuracy and the number of selected activation values.}
         \label{fig:activationRestriction}
\end{figure}

To demonstrate the tradeoff between the number of selected weight values and
the inference accuracy, we used different thresholds to select weight values
according to their power consumption and evaluated the accuracy by restricting
the neural networks to these weight values.  Figure~\ref{fig:spaceRestriction}
illustrates the results.  As expected, a lower power threshold leads to a lower
inference accuracy.  However, there is still a good potential for power
reduction before significant accuracy degradation appears.
For example, for
ResNet-50-CIFAR-100 the power threshold can be lowered down to 800\,$\mu$W,
which corresponds to 36 weight values,
leading to total power savings of 22.8\% with only a negligible accuracy loss.



Figure~\ref{fig:activationRestriction} shows the tradeoff between accuracy and
the number of activation values. The results are obtained by restricting neural
networks with different numbers of activation values based on a weight selection
threshold 825\,$\mu$W for LeNet5, ResNet-20 and ResNet-50, and a higher threshold 900\,$\mu$W for EfficientNet-B0-Lite.  The different numbers of activation values reflect
different maximum delays on the MAC unit.  In this figure, the left most point
corresponds to the full activation space with 256 activation values.  As the
number of activation values and thus the maximum delay decreases, the inference
accuracy is first well-maintained and then drops.  Before the turning point,
there is optimization potential we took advantage of  to enhance computational
performance or reduce power consumption by voltage scaling.


\section{Conclusion}
\label{sec:conclusion}

In this paper, we have proposed PowerPruning, a novel method to reduce power
consumption in digital neural network accelerators by selecting weights that
lead to less power consumption in MAC operations. The timing characteristics of
the selected weights together with activation transitions are also
evaluated.  We then selected weights and activations that lead to small delays,
so that either the clock frequency of the MAC units can be improved or voltage
scaling can be applied to reduce power consumption further.  Together with
retraining, the proposed method can reduce power consumption of DNNs on
hardware by up to 73.9\% with only a slight accuracy loss.  The proposed
method does not modify MAC units and can be combined
seamlessly with existing hardware architectures for power-efficient neural
network acceleration.

\section*{Acknowledgment} This work is funded by the Deutsche Forschungsgemeinschaft (DFG, German Research Foundation) – Project-ID 504518248 and by the National Science Foundation (NSF) 2112562.

\bibliographystyle{IEEEtran}
\bibliography{IEEEabrv,CONFabrv,bibfile}

\end{document}